\documentclass[journal]{IEEEtai}

\usepackage[table]{xcolor}
\usepackage{blindtext}
\usepackage{times}
\usepackage{epsfig}
\usepackage{amssymb}
\usepackage{amsthm}
\usepackage{graphicx}
\usepackage{amsfonts}
\usepackage{amsmath}
\usepackage{array}
\usepackage{mdwmath}
\usepackage{mdwtab}
\usepackage{eqparbox}
\usepackage{fixltx2e}
\usepackage{stfloats}
\usepackage{url}
\usepackage{diagbox}
\usepackage{verbatim}
\usepackage{booktabs}
\usepackage{multirow}
\usepackage{bm}
\usepackage{mathtools}
\usepackage{breqn}
\usepackage{float}
\usepackage{tcolorbox}
\usepackage{todonotes}
\usepackage{pgfmath}
\usepackage{siunitx}
\usepackage{adjustbox}
\usepackage{threeparttable}
\usepackage{hyperref}
\usepackage{flushend}
\usepackage{tabularx}
\usepackage{makecell}
\usepackage{pifont}
\usepackage{cite}
\usepackage{listings}
\usepackage[font=footnotesize,labelfont=bf]{caption}
\usepackage{subcaption}

\def\etal{{\it et al. }}

\def\ourmethod{MAPE}
\def\ourmodel{ShadowMaskFormer}

\definecolor{bb}{rgb}{0.0, 0.0, 0.5}
\def\etal{\emph{et al.}}

\newcommand{\revision}[1]{{\textcolor{black}{#1}}}
\newcommand{\revisiont}[1]{{\textcolor{red}{#1}}}

\usepackage{algorithmic}
\usepackage[ruled, vlined, linesnumbered]{algorithm2e}

\begin{document}

\title{\ourmodel{}: Mask Augmented Patch Embedding for Shadow Removal}


\author{Zhuohao Li, 
        Guoyang Xie$^{*}$,
        Guannan Jiang, 
        and~Zhichao Lu

\thanks{Zhuohao Li was with the Department of Computer Science, City University of Hong Kong, Hong Kong, China, and is currectly with the School of Ocean Engineering and Technology, Sun Yat-Sen University, Zhuhai 519082, China. Email: lizhh9810@gmail.com.}  
\thanks{Guoyang Xie was with the Department of Computer Science, City University of Hong Kong, Hong Kong, China, and is currently with the Department of Intelligent Manufacturing, CATL, Ningde 352000, China. Email: guoyang.xie@ieee.org. (\emph{Corresponding author: Guoyang Xie})}  
\thanks{Guannan Jiang is with the Department of Intelligent Manufacturing, CATL, Ningde 352000, China. Email: jianggn@catl.com.}  
\thanks{Zhichao Lu is with the Department of Computer Science, City University of Hong Kong, Hong Kong, China. Email: zhichao.lu@cityu.edu.hk.}
}

\markboth{IEEE Transactions on Artificial Intelligence, Vol. XX, No. XX, XXXX 2022}
{Zhuohao Li \MakeLowercase{\textit{et al.}}: \ourmodel{}: Mask Augmented Patch Embedding for Shadow Removal}

\maketitle

\begin{abstract}
Transformer recently emerged as the de facto model for computer vision tasks and has also been successfully applied to shadow removal. However, these existing methods heavily rely on intricate modifications to the attention mechanisms within the transformer blocks while using a generic patch embedding. As a result, it often leads to complex architectural designs requiring additional computation resources. 
In this work, we aim to explore the efficacy of incorporating shadow information within the early processing stage. Accordingly, we propose a transformer-based framework with a novel patch embedding that is tailored for shadow removal, dubbed \ourmodel{}. Specifically, we present a simple and effective mask-augmented patch embedding to integrate shadow information and promote the model's emphasis on acquiring knowledge for shadow regions. Extensive experiments conducted on the ISTD, ISTD+, and SRD benchmark datasets demonstrate the efficacy of our method against state-of-the-art approaches while using fewer model parameters.
%
Our implementation is available at \url{https://github.com/lizhh268/ShadowMaskFormer}.
\end{abstract}

\begin{IEEEImpStatement}
    Our research introduces \ourmodel{}, a transformer-based framework designed to enhance shadow removal in images. This new approach simplifies the process and improves efficiency, requiring fewer model parameters compared to existing methods. Technologically, \ourmodel{} integrates shadow information early in the processing stage, enabling more accurate and less resource-intensive image analysis. This can lead to more cost-effective AI applications where computational resources or power efficiency is a concern.
    Economically, the reduction in computational demand may lower the barriers to implementing advanced image-processing technologies in consumer electronics and other devices. Socially, by improving the quality of shadow removal, our method could enhance the visual experience in applications such as digital photography and video, making these technologies more accessible and enjoyable for users.
    \ourmodel{} contributes to the ongoing development of AI in visual computing by offering a more streamlined approach to a common problem, potentially influencing future advancements in the field.
\end{IEEEImpStatement}


\IEEEpeerreviewmaketitle

\section{Introduction} \label{sec:intro}
\IEEEPARstart{D}{eep} learning-based approaches have been widely used for various computer vision tasks by performing excellent performance over traditional model-based approaches~\cite{tai-1,tai-2,tai-3,tai-4}.
Shadow removal is highly challenging as it involves restoring irregular shadow regions and has gradually emerged as the dominant paradigm in this field. 
Recently, deep learning-based approaches~\cite{DeshadowNet2017,fu2021autoexposure,Hu_2020_DSC,wang2017stacked,ARGAN} have gradually emerged as the dominant paradigm in this field.
It should be emphasized that shadow mask which can distinguish shadow regions from non-shadow regions in an image, has been widely employed in deep learning methods and proven to effectively assist in shadow removal task~\cite{wang2017stacked,cun2019ghostfree,Hu_2019_MAskShadowGan}.

\begin{figure*}[t]
\begin{subfigure}[b]{0.21\textwidth}
\centering
\includegraphics[width=\textwidth]{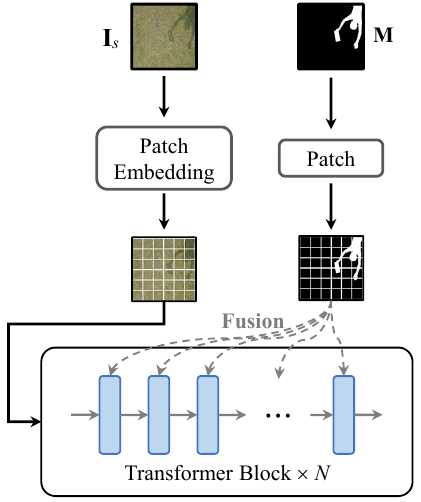}
\caption{Existing methods\label{fig:preview_existing}}
\end{subfigure} \hfill
\begin{subfigure}[b]{0.28\textwidth}
\centering
\includegraphics[width=\textwidth]{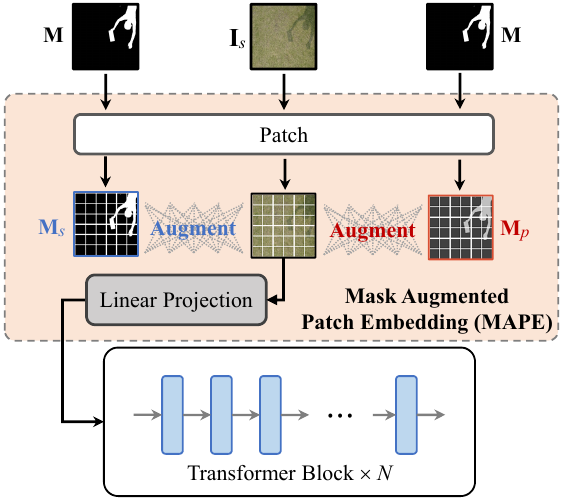}
\caption{ShadowMaskFormer\label{fig:preview_ours}}
\end{subfigure}\hfill
\begin{subfigure}[b]{0.50\textwidth}
\centering
\includegraphics[width=\textwidth]{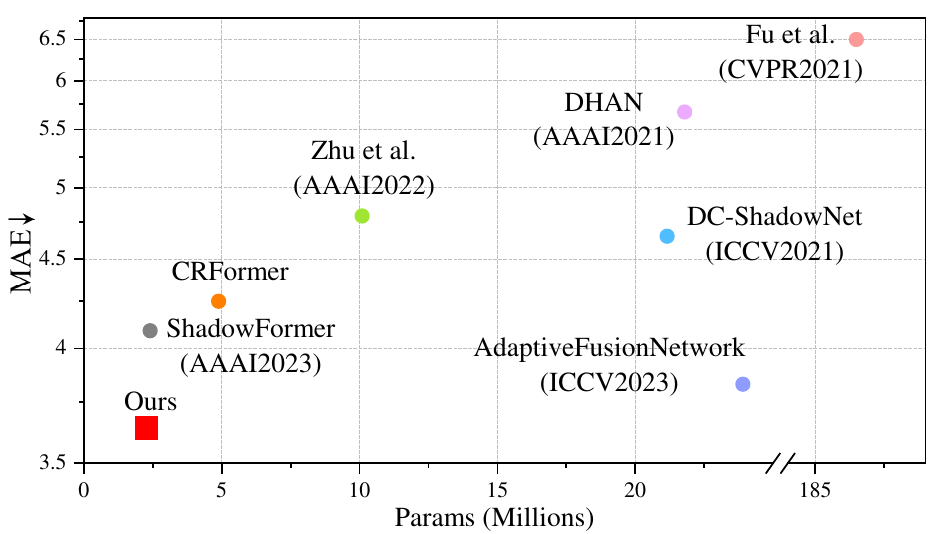}
\caption{Performance on the SRD dataset\label{fig:preview_ours_table}}
\end{subfigure}
\caption{(\emph{a}) Existing methods (such as CRFormer~\cite{wan2022crformer} and ShadowFormer~\cite{guo2023shadowformer} opt for vanilla patch embedding and focus on designing sophisticated modules to incorporate shadow information within the transformer blocks. In contrast, (\emph{b}) our method proposes to incorporate shadow information during the early processing stage and present a simple yet effective patch embedding module, dubbed \emph{\ourmethod{}}, tailored for shadow removal. (\emph{c}) Empirically, we demonstrate that our method leads to state-of-the-art performance on the SRD dataset with significantly lower computational complexity. For the shadow removal task, MAE is the mean absolute error computed in the LAB color space, where lower MAE indicates better performance. \label{fig:vs}} 
\end{figure*}

\vspace{2pt}
\noindent\textit{\textbf{Limitation of Shadow Transformer.}}
Recently, as an emerging backbone model of choice for vision tasks, transformers have also been applied to shadow removal tasks. However, 
there are two issues. \ding{182} The existing transformer-based methods overlook the shadow information in the early processing stage and directly employ the generic patch embedding, which cannot fully unleash the representational power of the transformer-based methods. As shown in Figure~\ref{fig:preview_existing}, the state-of-the-art transformer-based methods (i.e., CRFormer~\cite{wan2022crformer} and ShadowFormer~\cite{guo2023shadowformer}) achieve the contextual knowledge from the non-shadow regions, resulting in performance degradation. \ding{183} As shown in Figure~\ref{fig:preview_ours_table}, the transformer-based methods mainly necessitate the creation of new modules with shadow masks within the main computation blocks (i.e., transformer blocks and convolution blocks) for shadow removal, resulting in a noticeable scale of parameters. Hence, this has led to the following \emph{research question: Can we incorporate shadow information in patch embedding to avoid complicated modifications to transformer blocks and highlight the shadow region in the feature extraction stage? }

\vspace{2pt}
\noindent\textbf{\textit{\ourmodel{}.}}
To overcome the constraints of current transformer-based techniques and address the research question, we introduce a new framework called \ourmodel{}. As shown in Figure~\ref{fig:preview_ours}, \ourmodel{} combines the transformer model with a shadow mask in patch embedding, presenting a novel approach to effectively remove shadows from images. 
Specifically, in the early processing stage (i.e., patch embedding), we propose a simple and effective patch embedding, namely Mask Augmented Patch Embedding (\ourmethod{}).
The motivation behind \ourmethod{} is our observations based on the limited utilization of masks from existing work, as detailed in Section~\ref{sec:motivation}. 
Therefore, in our \ourmodel{}, the shadow mask is carefully utilized in \ourmethod{} with two complementary binarization schemes (the $0/1$ and $-1/+1$ Binarization) to enhance the shadow region pixels, as detailed in section~\ref{sec:mape}. 
In conclusion, with the proposed \ourmethod{}, our \ourmodel{} can not only leverage the contextual information that is achieved by the learning capability of the transformer model but also restore the shadow region pixels at an earlier time. 
This approach allows the model to more effectively acquire the distinctive features of shadow regions and perform shadow removal \revisiont{more purposefully}.
It is noteworthy that \ourmethod{} operates as a single-layer module, primarily performing pixel-level operations, and it requires computation only once per training epoch, significantly reducing computational demands.
As far as our knowledge extends, we believe our approach marks the pioneering exploration of utilizing patch embedding in vision transformer models for the task of shadow removal. 
Experimental results demonstrate that \ourmodel{} achieves outstanding shadow removal results over the three widely-used shadow removal datasets, surpassing the state-of-the-art performance.  Furthermore, compared with other SOTA methods, it only has 2.2MB network parameters, e.g., half of CRFormer~\cite{wan2022crformer} and smaller than others, as illustrated in Figure~\ref{fig:vs}. 
The main contributions of this work are as follows:


\begin{itemize}
\item[$\bullet$] \revision{We propose \ourmodel{}, a novel framework that integrates shadow mask information in the patch embedding stage. \emph{To the best of our knowledge, this is the first attempt to incorporate shadow-specific information at such an early stage in transformer models, establishing a new paradigm for shadow removal tasks.}}

\item[$\bullet$] \revision{We present \ourmethod{}, a \emph{simple yet effective method tailored for integrating shadow information, naturally aligned with the patch embedding stage}. By leveraging shadow masks with two complementary binarization schemes, \ourmethod{} enhances shadow region representations, enabling precise pixel-wise restoration without adding extra parameters. This results in both computational efficiency and robust performance across a wide range of scenarios.}

\item[$\bullet$]  \revision{Our method achieves state-of-the-art performance across benchmark datasets (ISTD, ISTD+, and SRD), demonstrating both exceptional accuracy and computational efficiency with a model size of only 2.2MB. Extensive experiments further demonstrate the robustness of MAPE even under incomplete or inaccurate shadow masks, proving its outstanding ability to generalize to complex, real-world conditions.}
\end{itemize}

The remainder of this paper is structured as follows:
Section II reviews related work in shadow removal and vision transformer.
Section III introduces the task background and the motivation of our proposed method.
Section IV describes the detailed architecture and workflow of \ourmodel{}.
Sections V and VI present our method's experimental results and ablation studies.
Finally, Section VII concludes our work and discusses future directions.

\section{Related Work\label{sec:related}}
In this section, We will present prior works related to image shadow removal, briefly outline vision transformers, and discuss their application in the task of shadow removal.
\subsection{Shadow Removal}
In the field of image processing, shadows pose a pervasive challenge that has negative implications for various downstream tasks, e.g., object detection, tracking, and face recognition~\cite{object,tracking,facerecognition}. Consequently, shadow removal has been a fundamental task in the field of computer vision and has received extensive research attention over the years. Numerous methods have emerged for shadow removal in images. These methods can be broadly categorized into two flavors: traditional model-based approaches and deep learning-based approaches. 
Traditional model-based approaches rely on the physical models of shadow images, which are limited by their dependence on prior knowledge and often struggle to effectively remove shadows in real-world scenes~\cite{Finlayson,prior2,priors}.
For instance,~\cite{Finlayson} proposed a physical model based on constant illumination conditions;~\cite{ShadowRemover} proposed an optimization model for removing shadow under various illumination conditions. 

In the recent past, deep learning-based methods have achieved excellent performance in the field of shadow removal, leveraging their end-to-end capabilities. 

Deep learning-based methods can be primarily categorized into two main approaches: convolutional neural network (CNN) based and generative modeling, particularly generative adversarial network (GAN), based approaches. 
CNN-based approaches employ multi-level convolutions to extract context-based features for shadow removal \cite{DeshadowNet2017,Hu_2020_DSC,le2019shadow,chen2021canet,zhu2022efficient,SG-ShadowNet,fu2021autoexposure,le2020ParamNet,XU2024109969}.
For instance, DeshadowNet~\cite{DeshadowNet2017} utilizes image contextual information to learn shadow mask features and remove shadows. 
GAN-based approaches generate shadow mask images and shadow-free images that closely resemble reality by adhering to a series of criteria for discerning between real and fake data \cite{ARGAN,Hu_2019_MAskShadowGan,wang2017stacked,cun2019ghostfree,Zhang2019RISGANER,jin2022dcshadownet,liu2021G2R,jin2022dcshadownet}.
For instance, in DC-ShadowNet~\cite{jin2022dcshadownet}, three dedicated losses were defined to characterize shadow images for accurate shadow synthesis. 
Besides, AdaptiveFusionNetwork~\cite{AF_Network_2023_ICCV} predicts the adaptive weights between two features that are extracted from the shadow image and the shadow mask image.
\revision{On the other hand, two recent works have made promising progress in shadow removal: ShadowDiffusion~\cite{guo2023shadowdiffusion} achieves notable performance by proposing a dynamic mask-aware diffusion model and incorporating degradation priors, while HomoFormer~\cite{xiao2024homoformer} offers another solution by using random shuffle and reverse shuffle operations to transform non-uniform shadows into a uniform distribution, enhancing the effectiveness of local self-attention for complex shadows.}
Our method is different from them, i.e., we define the problem as the integration of transformer models and shadow masks, which enables us to draw inspiration from existing mask utilization methods.

\subsection{Vision Transformer}
The transformer model was originally designed for natural language processing tasks~\cite{vaswani2017attention}, which were recently adopted for vision tasks.
Specifically, vision transformers segment images into patches and utilize them as inputs for the subsequent computation (i.e., multi-head attention) \cite{touvron2021training}.
Owing to the ability to learn global contextual information, transformers have achieved remarkable performance in vision tasks \cite{classformer,gberta_2021_ICML}.
%
%
Apart from regular image classification~\cite{classformer} and segmentation~\cite{segmenter2021}, transformer models have also been applied to various low-level vision tasks, such as image restoration~\cite{zamir2022restormer}, colorization~\cite{kumar2021colorization}, and inpainting~\cite{dong2022incremental}.
Differing from other image tasks, our objective is to propose a transformer-based framework for shadow removal. This allows us to design a module for utilizing shadow masks, achieving efficient shadow removal.
\subsection{Transformer for Shadow Removal}
%

With the remarkable performance of transformer models in computer vision tasks, it has also been applied to shadow removal tasks~\cite{wan2022crformer,guo2023shadowformer}.  CRFormer~\cite{wan2022crformer} attempts to guide the restoration of shadow regions by leveraging non-shadow region information with shadow masks. However, it places excessive emphasis on non-shadow region information, resulting in insufficient model attention allocation to the crucial shadow regions and shadow masks. 
ShadowFormer~\cite{guo2023shadowformer} proposes a Shadow-Interaction Module Attention to exploit the global contextual correlation between shadow and non-shadow regions. It is worth noting that the transformer-based method, still like other deep learning-based methods, relies on designing new modules from the transformer blocks to leverage shadow masks for the shadow removal task, as shown in Figure ~\ref{fig:vs}(a).
In this work, we aim to explore the efficacy of incorporating shadow information during the early processing stage and propose a novel patch embedding module tailored to the shadow removal task.
\section{Preliminaries \label{sec:Pre}}
In this section, we will provide the essential background knowledge about DL models and image shadow removal. Furthermore, we will elucidate the shadow model in this work and its motivation.

\subsection{Background}
In conventional computer vision models, such as convolutional neural networks, an input image is processed as an entirety by sliding a filter over the entire image to capture local patterns. 
In contrast, a vision transformer (ViT) model \cite{touvron2021training} does not directly operate on an input image, rather, it breaks down an input image into smaller and non-overlapping patches.
Each patch is then linearly transformed into a fixed-dimensional vector representation for subsequent computation. 
This process is known as \emph{Patch Embedding}.

The main computing backbone of a ViT model comprises $N$ sequentially connected transformer blocks. 
Each block comprises a multi-head attention (MHA), a feed-forward network (FFN), and LayerNorms \cite{ba2016layer}. 

%
\subsection{Motivation \label{sec:motivation}}
\noindent\textbf{Shadow masks.}
With a steady stream of promising empirical results confirming the effectiveness of shadow masks for shadow removal \cite{wang2017stacked,Hu_2019_MAskShadowGan,wan2022crformer,guo2023shadowformer}, we observe two patterns emerged from existing works: 
\ding{182} Shadow masks are typically represented as $0/1$ binary masks (i.e., $\textbf{M} \in \{0, 1\}^{H\times W}$ with {\small${H\times W}$} being the size of an input image) where ``$1$'' indicates shadow regions and ``$0$'' indicates non-shadow regions, respectively \cite{wang2017stacked,zhu2022efficient,wan2022crformer,guo2023shadowformer};
\ding{183} Shadow masks are primarily utilized by the main computing units of the models, e.g., convolution operations in CNN-based models \cite{maskshadownet,zhu2022efficient} and attention modules in ViT models \cite{wan2022crformer,guo2023shadowformer}.

Regarding observation \ding{182}, binarization to $0/1$ poses a potential risk of losing useful information as features corresponding to non-shadow regions will be completely suppressed if one directly applies $\textbf{M}$ to input signals. 
Instead of designing alternative ways to indirectly apply $\textbf{M}$ as done in most prior arts \cite{wang2017stacked,zhu2022efficient,wan2022crformer,guo2023shadowformer}, we opt for designing two complementary binarization schemes for direct utilization of $\textbf{M}$ in this work. 

In response to observation \ding{183}, we re-visit the location within a deep learning model where $\textbf{M}$ should be incorporated.
Specifically, we seek to investigate the efficacy of incorporating $\textbf{M}$ during the input preprocessing stage, i.e., the patch-embedding stage for ViT models, 
eliminating the need for repeated applications of $\textbf{M}$ in every transformer block, 
which in turn leads to improved model efficiency. 

\noindent\textbf{Physical models of shadow removal.}
For the model of shadow removal, based on the preceding works~\cite{4587491, SHEN20082461,physicalmodel}, it can be deduced that the formation of shadows occurs due to obstruction of direct illumination and a portion of the ambient illumination. This implies that the shadowed pixel ${I}_{x}^{shadow}$ will exhibit diminished intensity compared to their corresponding shadow-free pixels ${I}_{x}^{shadow-free}$. 
According to \cite{le2019shadow,physicalmodel}, we start from the original shadow illumination model, which describes a mapping function $T$ that transforms a shadow pixel $I^{shadow}_x$ to its non-shadow pixel $I^{shadow-free}_x$.
This mapping can be summarized as a linear function and the intensity of a lit pixel is formulated as:
\begin{equation}
\begin{aligned}
    I^{shadow-free}_x(\lambda) = L^d_x(\lambda)R_x(\lambda) + L^a_x(\lambda)R_x(\lambda)
\end{aligned}
\label{eqs1}
\end{equation}
where $\lambda$ is the wavelength and $I^{shadow-free}_x$ is the intensity reflected from an image pixel, $L$ and $R$ are the illumination and reflectance respectively, $L^d$ and $L^a$ denote the direct illumination and the ambient illumination, respectively.
A more detailed implementation of the shadow physics model can be found in the Appendix.
For a real shadow scene, an occluder blocks the direct illumination $L^d$ and part of the ambient illumination $L^a$, thus the shadowed pixel can be represented as:
\begin{equation}
\begin{aligned}
    I^{shadow}_x(\lambda) = a_x(\lambda)L^a_x(\lambda)R_x(\lambda)
\end{aligned}
\label{eqs2}
\end{equation}
where $a_x(\lambda)$ is the attenuation factor indicating the remaining fraction of $L^a$ that arrives at an image point $x$.

From these Equations, the shadow-free pixel can also be expressed as follows:
\begin{equation}\small
\begin{aligned}
    I^{shadow-free}_x(\lambda) = L^d_x(\lambda)R_x(\lambda) + a_x(\lambda)^{-1}I^{shadow}_x(\lambda)
\end{aligned}
\label{eqs3}
\end{equation}

Furthermore, following~\cite{le2019shadow}, we can establish a mapping function between ${I}_{x}^{shadow}$ and ${I}_{x}^{shadow-free}$. 
This mapping expresses the shadow-free pixel as a linear function of the shadowed pixel.
\begin{equation}
\begin{aligned}
    {I}_{x}^{shadow-free}(k) = w(k) \times {I}_{x}^{shadow}(k) + b_k
\end{aligned}
\label{eq-linear}
\end{equation}
where $k$ represents the color channel ($k$ $\in$ $R$,$G$,$B$), $b_k$ is the response of the camera to direct illumination, and $w(k)$ is responsible for the attenuation factor of the ambient illumination at this pixel in this color channel. Additionally, $w =[w_R,w_G,w_B]$ and $b = [b_R,b_G,b_B]$ are constant across all pixels $x$ in the umbra area of the shadow.
The crucial aspect of the learned mapping by the model lies in determining the parameters $w$ and $b$ for individual shadowed pixels. 
Inspired by this, we further assume the expression of Eq.~\ref{eq-linear} can be reformulated as follows:
\begin{equation}
\begin{aligned}
    {I}_{x}^{shadow-free}(k) = S(k) \times {I}_{x}^{shadow}(k) \nonumber
\end{aligned}
\label{eq-represented}
\end{equation}
In the above expression, the gain factor $S(k)$ in our method is derived from the assumption underlying Eq. 1, and $S(k)$ will be learned by the model for the shadowed pixel ${I}_{x}^{shadow}(k)$.
Specifically, considering the natural attribute of shadows where the attenuation of light sources results in shadow pixel values significantly lower than those of non-shadowed pixels, we assume that in Eq. 1, $w(k)$ will be noticeably greater than 1, and $b_k$ will be smaller compared to ${I}_{x}^{shadow}(k) \times w(k)$. 
The term ${I}_{x}^{shadow}(k)$ appears to be roughly equivalent to the body reflection.
Hence, the crux of shadow removal lies in ensuring that the gain factor $S(k)$ learned by the model closely approximates the correct solution. In this paper, we endeavor to perform a preliminary exploration of the gain factor $S(k)$ during the patch embedding stage. This approach allows the model to enhance shadow removal performance more effectively at an early stage, thereby reducing unnecessary model exploration. Specifically, to address the task of shadow removal, we introduce a simple and effective patch embedding, namely Mask Augmented Patch Embedding.

\begin{figure*}[t]
\centering
\includegraphics[width=.9\textwidth]{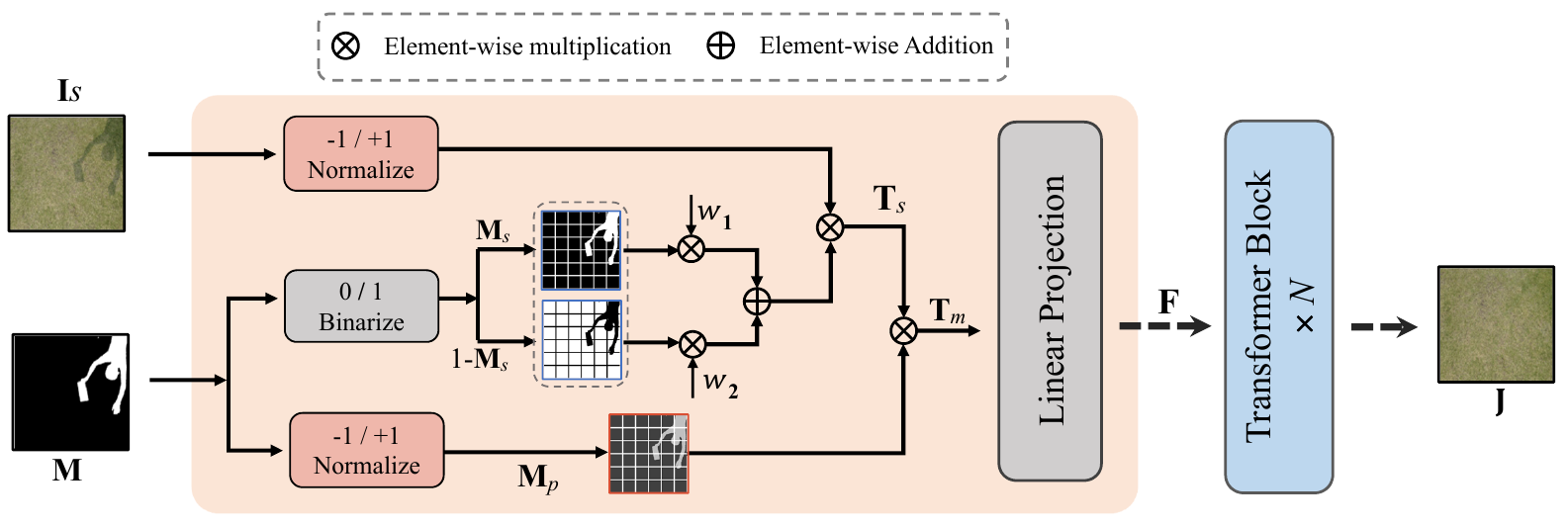} 
\caption{An Overview of our \ourmodel{} framework with the proposed mask augmented patch embedding (\ourmethod{}). First, MAPE takes the shadow image $\textbf{I}_s$ and its corresponding shadow mask \textbf{M} as inputs. Then, different processing techniques are applied to generate refined shadow mask $\textbf{M}_s$ and $\textbf{M}_p$ and to enhance the shadow region pixels. Subsequently, N transformer Blocks learn contextual information from MAPE.}
\label{fig:stru_res} 
\end{figure*}

\begin{algorithm}[tb]
\caption{Mask Augmented Patch Embedding (\ourmethod{})}
\label{alg:code}
\definecolor{codeblue}{rgb}{0.25,0.5,0.5}
\definecolor{codekw}{rgb}{0.85, 0.18, 0.50}
\begin{lstlisting}[language=python, mathescape, numbers=none, basicstyle=\linespread{1.1}\ttfamily, aboveskip=0pt, belowskip=0pt]
# x: shadow images
# mask: shadow mask images
# w1, w2: weight factors
# proj: Linear Projection
class MAPE(nn.Module):
    def __init__(self, patch_size, 
        in_chans, embed_dim, kernel_Size):
        super(MAPE, self).__init__()
        self.patch_embed = PatchEmbed(
        patch_size, in_chans, embed_dim, 
        kernel_size)
        
    def forward(self, x, mask, w1, w2):
        x = (x / 255) * 2 - 1 
        Ms = mask / 255
        Ts = (Ms * w1 + (1 - Ms) * w2) * x 
    # the first step of mask angmented
        Mp = (mask / 255) * 2 - 1 
        Tm = Mp * Ts
        F = self.proj(Tm) 
    # the second step of mask angmented
        return F
\end{lstlisting}

\end{algorithm}

\section{\ourmodel{}}
In this section, we will elucidate the overarching workflow framework of the model, with a particular emphasis on detailing the implementation of Mask Augmented Patch Embedding.

\subsection{Overview}
An overview of the proposed transformer with mask augmented in the patch embedding stage (\ourmodel{}) is depicted in Figure~\ref{fig:stru_res}. This approach employs two types of shadow masks to enhance the shadow region pixels. To achieve this, in the early processing stage of the model, \ourmodel{} integrates a Mask Augmented Patch Embedding (MAPE) specifically designed for shadow removal.
\par
Specifically, in the patch embedding stage, MAPE takes the shadow image $\textbf{I}_s$ and its corresponding shadow mask \textbf{M} as inputs. The main idea behind MAPE is to enhance the shadow region pixels in the early stage by utilizing the information of the $\textbf{M}$ and $\textbf{I}_s$. 
In \ourmodel{}, the determination of $S(k)$ is achieved through two forms of mask utilization: initialization at the early stage and adaptive refinement through model exploration.
After MAPE, the transformer blocks, based on the vision transformer, leverage its powerful ability to learn contextual information. It aims to learn a nonlinear mapping function $f(\textbf{I}_s, \textbf{M}; \theta)$ from $\textbf{I}_s$ to $\textbf{I}_{gt}$. By applying this mapping function, the model reconstructs the shadow-free image $\textbf{J}$ which represents the image after shadow removal.
\par

\subsection{Mask Augmented Patch Embedding \label{sec:mape}}
Based on the discussions above, aiming to find an accurate approximation of the gain factor $S(k)$ for shadow region pixels early in the model's training, we introduce the concept of Mask Augment\revisiont{ed} Patch Embedding which is based on the two complementary binarization schemes (the $0/1$  and  $-1/+1$ Binarization) as depicted in Figure~\ref{fig:stru_res}.
In particular, to ensure the learned non-linear mapping function $f(\textbf{I}_s, \textbf{M}; \theta)$ of the model to explore the correct $S(k)$ early and avoid exploring useless information for shadow region enhancement, we designed MAPE to be positioned in the early stage of the model. In the patch embedding stage, the shadow image $\textbf{I}_s$ and its corresponding shadow mask $\textbf{M}$ are first used as inputs to the model. 
Subsequently,  the $0/1$  and  $-1/+1$ (Binarize) pixel-wise operations are performed on $\textbf{M}$, resulting in two complementary masks $\textbf{M}_s$ and $\textbf{M}_p$, respectively.
These operations can be expressed using the following formulas:
\begin{equation}
\begin{aligned}
\textbf{M}_{s} = (\frac{\textbf{M}} {255})
\end{aligned}
\label{eq2}
\end{equation}

\begin{equation}
\begin{aligned}
    \textbf{M}_{p} = (\frac{\textbf{M}} {255}) \cdot 2 - 1
\end{aligned}
\label{eq3}
\end{equation}
After applying binarization, pixels corresponding to the shadow regions of $\textbf{M}_s$ are set to 1, while the non-shadow regions are set to 0. $\textbf{M}_p$ undergoes pixel-wise operations to adjust the pixel distribution to the range [-1, 1].
\par
First, we use a set of $0/1$ masks $\textbf{M}_s$ to learn weighted masks for adjusting the intensity of pixels from the shadow (S) and non-shadow (NS) regions, which in essence provides an initial estimation of $S(k)$.
Thus we can obtain the first enhanced shadow image $\textbf{T}_s$. Subsequently, the region information carried by $\textbf{M}_p$ is combined with $\textbf{T}_s$ through further operations, yielding the feature $\textbf{T}_m$ with element-wise multiplication. Finally, we employ linear projection on the $\textbf{T}_m$, introducing linear transformation features. These operations of mask-augmented patch embedding can be expressed as follows.
\begin{equation}
\begin{aligned}
    \textbf{T}_{s} = (w1 \cdot \textbf{M}_{s} + w2 \cdot (\textbf{1} -\textbf{M}_{s})) \cdot \textbf{I}_{s}
\end{aligned}
\label{eq4}
\end{equation}

\begin{equation}
\begin{aligned}
    \textbf{T}_{m} = \textbf{M}_{p} \cdot \textbf{T}_{s}
\end{aligned}
\label{eq41}
\end{equation}

\begin{equation}
\begin{aligned}
    \textbf{F} = LinearProjection(\textbf{T}_{s})
\end{aligned}
\label{eq42}
\end{equation}
where $w1$ and $w2$ are the weights for region reassignment. $\textbf{F}$ is the output of MAPE, and $LinearProjection$  denotes the convolutional layer with the kernel size of 3 × 3.  Note that due to performance and efficiency by introducing convolutions into ViT~\cite{convproj}, we opted for convolutional projection rather than positional projection.

Actually, the process of mask augmented can be seen as two steps and its pseudocode can be summarized as Algorithm~\ref{alg:code}. 
In the first step, we reassign the pixel information in the shadow and non-shadow regions by using the region reassignment weights $w1$ and $w2$ with $0/1$ mask, respectively. This can be expressed as $(w1 \cdot \textbf{M}_{s} + w2 \cdot (1-\textbf{M}_{s})) \cdot \textbf{I}_{s}$, which means $S(k) = w1$.
It should be noted that in order to enhance the shadow region pixels, it is necessary to set $w1$ greater than $w2$. In our experiments, we empirically set $w1$ and $w2$ to 2.5 and 1, respectively.
Besides, we conducted a feature analysis of shadow images, e.g.: For ISTD datasets, the results indicate that after the $-1/+1$ processing, the proportion of positive values in the non-shadow regions reaches 70\%. In contrast, in the shadow regions, negative values overwhelmingly dominate, reaching 97\% of the total values.
As shadow region pixel values tend to be low, the majority of shadow region pixels of $\textbf{T}_s$ are negative and this contradicts with the pixels in the non-shadow region (positive).
Therefore, the second step is introduced to further enhance shadow region pixels by adaptive refinement. 
We use $-1/+1$ masks to re-balance the distribution of the normalized pixel intensity from the shadow and non-shadow regions, where most shadow pixels are negative-valued while most non-shadow pixels are positive-valued. This step essentially aims to make the restored shadow regions closely resemble non-shadow regions. 
Finally, we applied linear projection to $\textbf{T}_m$, transforming it into the input for the transformer block. At this point, $S(k) = LinearProjection(w1; k)$. Note that the transformer blocks also simultaneously explore the genuine $S(k)$ values for each pixel during the training phase.
In addition, MAPE avoids intricate modifications to the transformer blocks and eliminates the involvement of shadow masks in the computation of every transformer block, leading to noticeable computational savings.

\subsection{Loss Function}
Following prior shadow removal works \cite{le2019shadow,SG-ShadowNet,wan2022crformer}, we only adopt the image consistency loss $\mathcal{L}1$, which is mathematically defined as follows:
\begin{equation}
\begin{aligned}
    \mathcal{L}1 = ||\ \mathbf{J} - \mathbf{I}_{gt}\ ||_{1}
\end{aligned}
\end{equation}
where $\mathbf{J}$ and $\mathbf{I}_{gt}$ are the predicted shadow-free image and the ground truth, respectively.
\revision{The use of $\mathcal{L}1$ loss is motivated by its ability to effectively capture pixel-wise differences, which is crucial for accurately removing shadows while preserving fine details. This makes $\mathcal{L}1$ loss well-suited to meet the requirements of our proposed framework.}
\section{Experiments}\label{exper}
This section briefly describes our experimental setup followed by results on multiple shadow removal benchmarks. 

\begin{table}[t]
    \centering
    \caption{
    The MAE results of shadow regions (\textbf{S}), non-shadow regions (\textbf{NS}), and whole images (\textbf{All}) on the ISTD dataset. The best and the second-best results in each section are highlighted in bold and underlined, respectively. $^{\dagger}$ indicates that the results are evaluated under input size of 400 $\times$ 400.\label{tab:istd}}
    \renewcommand{\arraystretch}{1.2}
    \setlength{\tabcolsep}{0.4mm}{
    \resizebox{0.9\linewidth}{!}{
    	{\begin{tabular}{c|l|c|c|c|c}
    		\toprule[1.0pt]
    		 & \multirow{2}{*}{Method} & \multirow{1}{*}{Params} &\multicolumn{1}{c|}{\textbf{S} }&\multicolumn{1}{c|}{\textbf{NS}}& \multicolumn{1}{c}{\textbf{All}}\\ 
    		 & & (M) &MAE$\downarrow$ &MAE$\downarrow$ &MAE$\downarrow$ \\ 
    		\midrule[0.5pt]
    		\multirow{13}{*}{\rotatebox{90}{256 $\times$ 256}}
                & Input Image & - & 32.1 & 7.09 & 10.9\\
                  & Guo \etal~\cite{Guo2012} & - & 18.7 & 7.76 & 9.26 \\
                  &  MaskShadow-GAN~\cite{Hu_2019_MAskShadowGan}  & 13.8 & 12.7 & 6.68 & 7.41\\
                   & ST-CGAN~\cite{wang2017stacked} & 31.8 & 9.99 & 6.05 & 6.65\\
                   & DSC~\cite{Hu_2020_DSC} & 22.3 & 8.72  & 5.04 & 5.59\\
                  &  G2R~\cite{liu2021G2R}  & 22.8 & 10.7 & 7.55 & 7.85\\
                   & DHAN~\cite{cun2019ghostfree}  & 21.8 & 7.49 & 5.30 & 5.66\\
                   & Fu~\etal~\cite{fu2021autoexposure} & 187 & 7.91 & 5.51 & 5.88\\
                   & DC-ShadowNet~\cite{jin2022dcshadownet}  & 21.2 & 11.4 & 5.81 & 6.57\\
                   & Zhu~\etal~\cite{zhu2022efficient} & 10.1 & 8.29 & 4.55 & 5.09\\
                   &  CRFormer$^{\dagger}$~\cite{wan2022crformer} & 4.89 & 7.32 & 5.82 & 6.07\\
               & ShadowDiffusion~\cite{guo2023shadowdiffusion} & 55.52 & \textbf{4.13} & 4.14 & \textbf{4.12} \\
               &  ShadowFormer~\cite{guo2023shadowformer} & 2.40 & 6.16 & \underline{3.90} & 4.27\\
               \cline{2-6}
                 & \revision{ShadowMaskFormer (ours)} & \textbf{2.28} & \underline{6.08} & \textbf{3.86} & \underline{4.23}  \\
    	    \midrule[0.5pt]
                \multirow{6}{*}{\rotatebox{90}{480 $\times$ 640}} 
                & Input Image & - & 33.23 & 7.25 & 11.4\\
                & ARGAN~\cite{ARGAN} & - & 9.21 & 6.27 & 6.63 \\
                & DHAN~\cite{cun2019ghostfree} & 21.8 & 8.13 & 5.94 & 6.29 \\
                & CANet~\cite{chen2021canet} & 358 & 8.86 & 6.07 & 6.15 \\
                & ShadowFormer~\cite{guo2023shadowformer} & 2.40 & 6.93 & 4.59 & 4.96 \\
                \cline{2-6}
                & ShadowMaskFormer (ours)  & \textbf{2.28} & \textbf{6.83} & \textbf{4.50} & \textbf{4.87} \\
                \bottomrule[1.0pt]
    \end{tabular}}}} 
\end{table}

\begin{table}[t]
    \centering
    \caption{The MAE results of shadow regions (\textbf{S}), non-shadow regions (\textbf{NS}), and whole images (\textbf{All}) on the ISTD+ dataset. The best and the second-best results in each section are highlighted in bold and underlined, respectively. $^\dag$ indicates that the results are evaluated under an input size of 400 $\times$ 400. \label{tab:istd+}}
    \renewcommand\arraystretch{1}
    \setlength{\tabcolsep}{2.0mm}{
    \resizebox{\linewidth}{!}{
    	{\begin{tabular}{l|c|c|c|c}
    		\toprule[1.0pt]
    	\multirow{2}{*}{Method} &\multirow{1}{*}{Params} & {\textbf{S} }& {\textbf{NS}}& {\textbf{All}}\\ 
    		&(M) &MAE$\downarrow$ &MAE$\downarrow$ &MAE$\downarrow$ \\ 
    		\midrule[0.5pt]
                     Input Image  &- & 40.2 & 2.6 & 8.5\\
                \midrule[0.5pt]
                     DHAN~\cite{cun2019ghostfree} &21.8  & 11.2   & 7.1  & 7.8\\
                     Param-Net~\cite{le2020ParamNet} &- & 9.7  & 2.9 & 4.1\\
                     G2R~\cite{liu2021G2R}   &22.8  & 8.8   & 2.9 & 3.9\\
                     SP+M-Net~\cite{le2019shadow} &141 & 7.9   & 2.8  & 3.6\\
                     Fu~\etal~\cite{fu2021autoexposure} &187  & 6.6   & 3.8  & 4.2\\
                SG-ShadowNet$^\dag$~\cite{SG-ShadowNet} &6.20 & 5.9 & 2.9 & 3.4 \\
               \revision{ShadowDiffusion}~\cite{guo2023shadowdiffusion} & 55.52 & \textbf{4.9} & \underline{2.3} & \textbf{2.7} \\
               ShadowFormer~\cite{guo2023shadowformer} &2.40  & \underline{5.4} & 2.4 & \underline{2.8}\\
               AdaptiveFusionNetwork~\cite{AF_Network_2023_ICCV} & 23.9 & 5.9 & 2.9 & 3.4 \\
               \midrule[0.5pt]
               ShadowMaskFormer (ours) &\textbf{2.28} & \underline{5.4} & \textbf{2.2} & \textbf{2.7} \\
    		\bottomrule[1.0pt]
    \end{tabular}}}}
\end{table}

\subsection{Experimental Setup} \label{sec:setup}
\noindent\textbf{Datasets.}
We evaluate our method on three widely-used benchmarks for shadow removal: 
(1) ISTD dataset \cite{wang2017stacked} comprises 1,870 image triplets (i.e., shadow images, shadow-free images, and shadow masks), from which 1,330 and 540 triplets are used for training and testing, respectively; 
(2) Adjusted ISTD (ISTD+) dataset \cite{le2019shadow} reduces illumination inconsistency between shadow and shadow-free images using an image processing algorithm while retaining the same number of triplets as the original ISTD dataset;
(3) SRD dataset \cite{DeshadowNet2017} comprises 2,680 training and 408 testing pairs of shadow and shadow-free images w/o ground truth shadow masks. Accordingly, we adopt the predicted masks provided by \cite{cun2019ghostfree} for training and evaluation. 

\vspace{3pt}
\noindent\textbf{Baselines.} Towards obtaining a representative evaluation of our method, we consider a wide range of baselines, including a traditional method \cite{Guo2012}, generative modeling-assisted approaches \cite{Hu_2019_MAskShadowGan,wang2017stacked,liu2021G2R,cun2019ghostfree,ARGAN,AF_Network_2023_ICCV}, two transformer-based approaches \cite{wan2022crformer,guo2023shadowformer}, among others \cite{Hu_2020_DSC,jin2022dcshadownet,chen2021canet, guo2023shadowdiffusion}.

\vspace{3pt}
\noindent\textbf{Evaluation metrics.} 
Following the previous methods~\cite{le2019shadow,le2020ParamNet,xiao2024homoformer}, we use mean absolute error (MAE) between the estimated shadow-free images $\mathbf{J}$ and the ground truth $\mathbf{I}_{gt}$ in the LAB color space, where lower values indicate better results.
We also use the peak signal-to-noise ratio (PSNR) and the structural similarity (SSIM) \cite{ssim} metrics to quantitatively compare the performance of various methods in the RGB color space, where higher values indicate better results.

\vspace{3pt}
\noindent\textbf{Implementation details.}
We implement our method on top of a recent image restoration framework \cite{DehazeFormer} with swin transformer \cite{SwinTransformer}. 
For training, we use AdamW \cite{AdamW} optimizer with a batch size of one and an initial learning rate of $2\times10^{-4}$, which is annealed to zero in 300 epochs following the cosine schedule \cite{loshchilov2017sgdr}. 
All experiments are carried out on two NVIDIA GeForce GTX 3090 GPUs. 

\begin{figure*}[t]
    \centering
    \includegraphics[width=.95\textwidth]{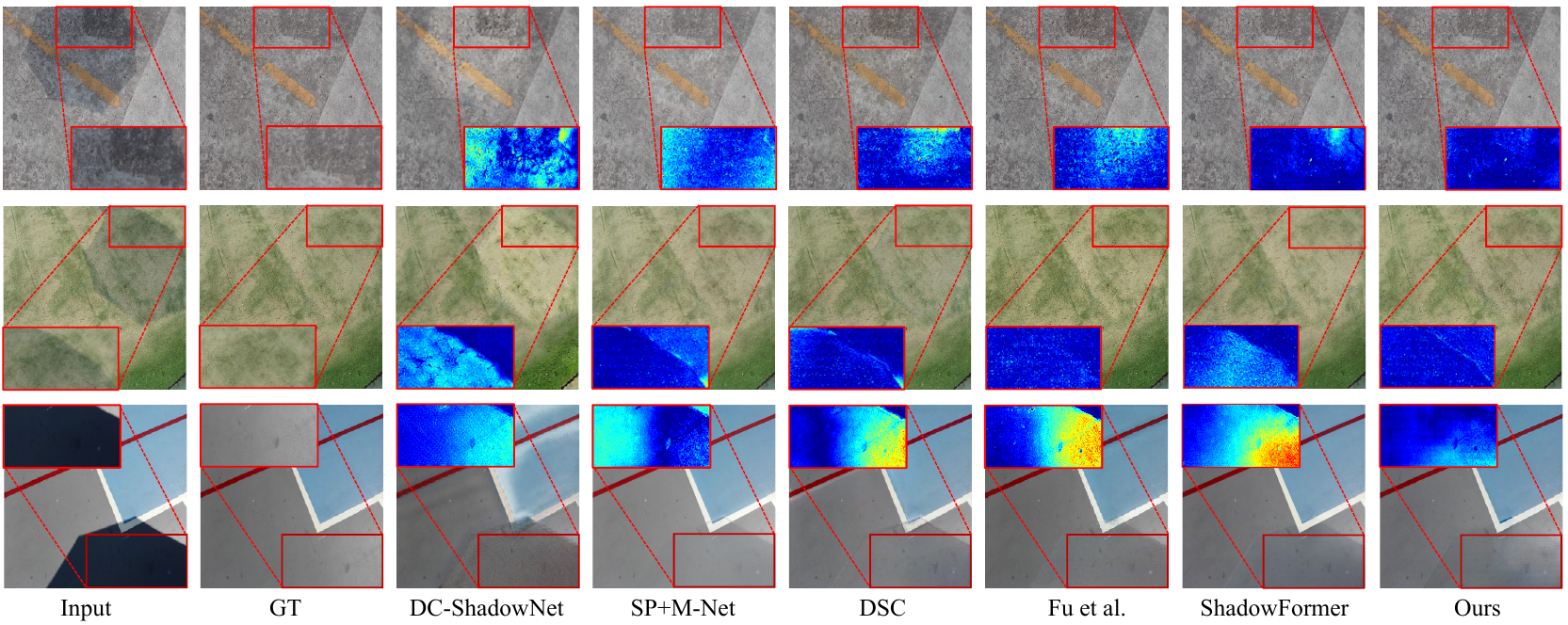} 
    \caption{Visual comparison among various approaches on examples from the ISTD dataset. We show the input image, the ground truth, and the results from DC-ShadowNet, SP+M-Net, DSC, Fu~\etal, \revision{ShadowFormer}, and our method, respectively from left to right. Heatmaps depict the differences between results and ground truth. Zoom in for details. \label{fig:istd_res}} 
\end{figure*}

\begin{table*}[t]
\centering
\caption{Comparison of shadow removal performance, measured by PSNR, SSIM, and MAE metrics, on the SRD dataset. The best and the second-best results in each section are highlighted in bold and underlined, respectively. ``-'' indicates that the information is not available publicly. \label{tab:srd}}
\setlength{\tabcolsep}{0.6em}
\renewcommand{\arraystretch}{1}
\adjustbox{width=0.9\linewidth}{
   \begin{tabular}{l|c | ccc| ccc| ccc}
       \toprule
                \multirow{2}{*}{Method}  & {Params} & \multicolumn{3}{c|}{Shadow Region (\textbf{S})}  &
                \multicolumn{3}{c|}{Non-Shadow Region (\textbf{NS})}  &
                \multicolumn{3}{c}{All Image (\textbf{All})} \\
                & (M) & PSNR$\uparrow$ & SSIM$\uparrow$ & MAE$\downarrow$ & PSNR$\uparrow$ & SSIM$\uparrow$ & MAE$\downarrow$ & PSNR$\uparrow$  & SSIM$\uparrow$ & MAE$\downarrow$ \\
                    \midrule[0.5pt]
                     Input Image  &- & 18.96 & 0.871  & 36.69 & 31.47 & 0.975  & 4.83 & 18.19 & 0.830  & 14.05\\
                     \midrule[0.5pt]
                     DeshadowNet~\cite{DeshadowNet2017} &- & - & - & 11.78 & - & - & 4.84 & - & - & 6.64\\
                     DSC~\cite{Hu_2020_DSC}  &22.3  & 30.65 & 0.960  & 8.62  & 31.94 & 0.956  & 4.41 & 27.76 & 0.903  & 5.71\\
                     DHAN~\cite{cun2019ghostfree} &21.8  & 33.67 & 0.978  & 8.94  & 34.79 & 0.979  & 4.80 & 30.51 & 0.949  & 5.67\\
                     Fu~\etal~\cite{fu2021autoexposure} &187 & 32.26 & 0.966  &9.55  & 31.87 & 0.945  & 5.74 & 28.40 & 0.893  & 6.50\\
                     DC-ShadowNet~\cite{jin2022dcshadownet} &21.2 & 34.00 & 0.975  &7.70  & 35.53 & 0.981  & 3.65 & 31.53 & 0.955  & 4.65 \\
                     CANet~\cite{chen2021canet} &358 & - & -  &7.82 & - & -  & 5.88 & - & -  & 5.98 \\
                     SG-shadowNet~\cite{SG-ShadowNet} &6.20 &- & - & 7.53 & - & - & \underline{2.99} & - & - & 4.23\\
              Zhu~\etal~\cite{zhu2022efficient} &10.1 & 34.94 & 0.980  & 7.44  & 35.85 & 0.982 & 3.74  & 31.72 & 0.952  & 4.79 \\
              CRFormer~\cite{wan2022crformer} &4.89 & - & -  & 7.14 & - & - & 3.15 & - & -  & 4.25\\
              ShadowDiffusion~\cite{guo2023shadowdiffusion} & 55.52 & \textbf{38.72} & 0.987 & \textbf{4.98} & 37.78 & \underline{0.985} & 3.44 & \textbf{34.73} & \textbf{0.970} & \textbf{3.63}\\
              ShadowFormer~\cite{guo2023shadowformer} &2.40 & 36.13 & 0.988 & 6.05 & 35.95 & \textbf{0.986} & 3.55 & 32.38 & 0.955 & 4.09\\
              \midrule[0.5pt]
              ShadowMaskFormer (ours) & \textbf{2.28} & \underline{37.71} & \textbf{0.988}  & \underline{5.55} & \textbf{38.23} & 0.984  & \textbf{2.98} & \underline{34.43} & \underline{0.968}  &\underline{3.64} \\
             \bottomrule
       \end{tabular}
   }
\end{table*}

\subsection{Experimental Results}
\noindent\textbf{Results on ISTD.}
Table~\ref{tab:istd} summarizes the results on the ISTD dataset. 
In general, we observe that our method consistently outperforms all considered baselines on different regions (i.e., shadow and non-shadow regions) under both input resolution settings (i.e., $256\times 256$ and $480\times 640$) with fewer model parameters. 
In particular, our method achieves \emph{1.6${\times}$ lower MAE} (averaged over whole images) with \emph{9.3${\times}$ fewer parameters} than DC-ShadowNet \cite{jin2022dcshadownet}.
Our method also achieves \emph{1.4${\times}$ lower MAE} (averaged over whole images) with \emph{2.1${\times}$ fewer parameters} than CRFormer \cite{wan2022crformer}.
Our method is also all-around competitive against ShadowFormer \cite{guo2023shadowformer}. 

Additional qualitative comparisons are provided in Figure~\ref{fig:istd_res}. In general, the empirical improvements provided by our method (as depicted in Table~\ref{tab:istd}) translate well to visualization results.
Evidently, we observe that our method significantly outperforms existing approaches in removing shadows from complex scenes (i.e., the second row), achieving a more natural adaptation between shadow and non-shadow regions.
For example, SP+M-Net~\cite{le2019shadow} tends to incorrectly process the non-shadow regions excessively, e.g., the results of the second and third rows in the fourth column.
DC-ShadowNet~\cite{jin2022dcshadownet} fails to successfully remove shadows and also affects the non-shadow regions, as shown in the result of the third row in the third column.

\begin{figure*}[t]
    \centering
    \includegraphics[width=.94\textwidth]{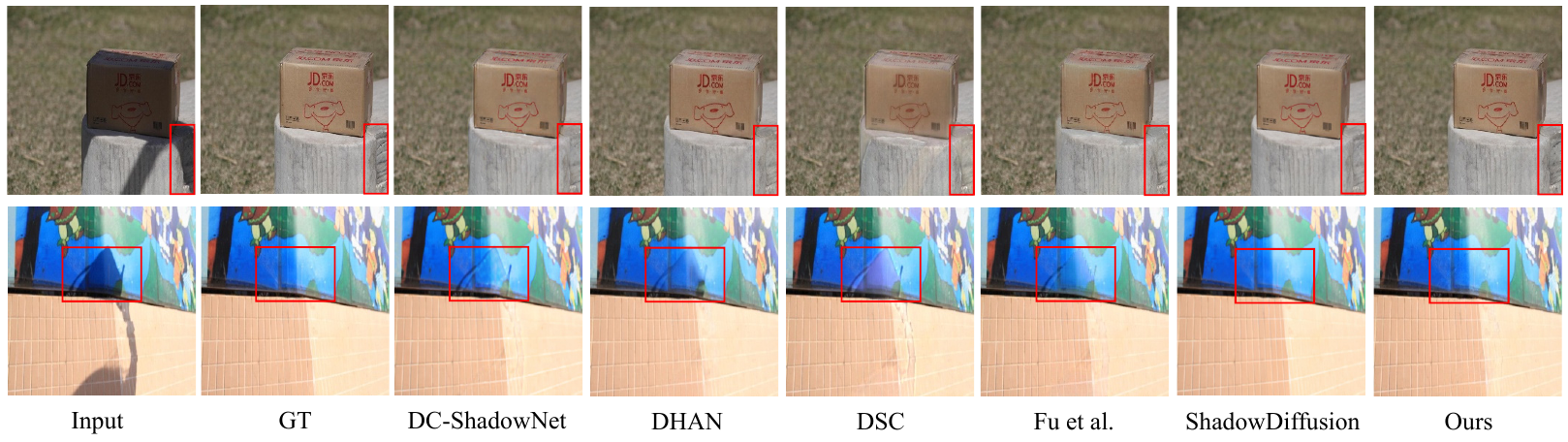} 
    \caption{Visual comparison among various approaches on examples from the SRD dataset. We show the input image, the ground truth, and the results from DC-ShadowNet, DHAN, DSC, Fu~\etal, \revision{ShadowDiffusion}, and our method, respectively from left to right. 
    \label{fig:srd_res} 
    }
\end{figure*}

\vspace{3pt}
\noindent\textbf{Results on ISTD+.}
Table~\ref{tab:istd+} summarizes the results on the ISTD+ dataset.
In general, we observe that (i) all methods exhibit better performance than that on the original ISTD dataset \cite{wang2017stacked} after adjustments of illumination inconsistency introduced by \cite{le2019shadow}; 
(ii) our method continues to outperform all the considered baseline competitors as indicated by MAE while using fewer model parameters. 
Specifically, our method achieves better MAE on shadow, non-shadow, and all regions than both SG-ShadowNet \cite{SG-ShadowNet} and ShadowFormer \cite{guo2023shadowformer}. 

\vspace{3pt}
\noindent\textbf{Results on SRD.}
Table~\ref{tab:srd} presents the comparison results on the SRD~\cite{DeshadowNet2017} dataset. 
Evidently, we observe that our method achieves significantly better results on the overall image than all considered baselines.
For instance, our method yields 1.58dB to 2.28dB PSNR improvements on shadow, non-shadow, and all regions over ShadowFormer \cite{guo2023shadowformer} with a similar number of model parameters. 
In addition, we observe that our method also achieves noticeably better MAE than the baselines. 
Specifically, our method achieves 1.3$\times$ and 1.2$\times$ lower MAE than CRFormer \cite{wan2022crformer} on shadow and all regions, respectively, while using 2.1$\times$ fewer model parameters.

Figure~\ref{fig:srd_res} depicts qualitative comparisons between our method and baselines. 
Evidently, we observe that our method not only can better eliminate artifacts while removing shadows but also effectively restore the content and color of the shadow regions.

\section{Ablation Studies\label{sec:ablation}}
In this section, we conduct a detailed analysis of each component of our \ourmethod{} framework, with the number of transformer blocks N set to 5. 
We conduct ablation studies on the ISTD dataset~\cite{wang2017stacked} to evaluate the individual impact of these components. In our paper, we argue that the mask deep fusion approach efficiently utilizes shadow mask and provides useful and sufficient information about shadow regions.

\vspace{3pt}
\noindent\textbf{Ablation Study on \ourmethod{}.}
%
As summarized in Table \ref{tab:ablation}, we first present the performance of the regular path embedding \cite{dosovitskiy2021an}, which is the de-facto choice of image preprocessing adopted by most existing works that are based on transformers.
Evidently, our proposed mask augmented patch embedding leads to significantly better shadow removal performance measured by MAE.
Compared to the conventional patch embedding, our MAPE can obviously utilize the regional information provided by the shadow mask effectively and convey this information to the entire model.
%
Then, we present the relative improvements from the proposed $-1/+1$ binarization scheme by replacing it with the regular 0/1 binary mask (i.e., $\textbf{M}_{p}$ $\xrightarrow{}$ $\textbf{M}_{s}$), where the effectiveness of $\textbf{M}_{p}$ can be observed.
This experimental result demonstrates that the combination of the two shadow masks designed in our MAPE is significantly more effective in shadow removal compared to using only a single, common 0/1 mask ($\textbf{M}_{s}$). 
In other words, the success of MAPE relies on the reliable combined use of the two masks. On the contrary, applying shadow masks directly within patch embedding can lead to performance degradation in shadow regions, as shown in Table~\ref{tab:ablation}.

\begin{table}[ht]
\centering
\caption{Comparison among the de-facto patch embedding (PE), our proposed \ourmethod{}, and a variant of \ourmethod{} on the ISTD dataset. Relative differences are shown in parentheses. \label{tab:ablation}}
\setlength{\tabcolsep}{0.4em}
\renewcommand{\arraystretch}{0.7}
\adjustbox{width=0.9 \linewidth}{
    \begin{tabular}{l|c|c|c}
        \toprule
         \multirow{2}{*}{Method} & {S}  & {NS}& {All}\\
         & MAE$\downarrow$  & MAE$\downarrow$ & MAE$\downarrow$ \\
                \midrule
                Original PE & 7.59 (\textcolor{red}{+1.5}) & 4.91 (\textcolor{red}{+1.1}) & 5.35 (\textcolor{red}{+1.1})\\\midrule
                \ourmethod{} & \textbf{6.08} (+0.0)  & \textbf{3.86} (+0.0) & \textbf{4.23} (+0.0) \\\midrule
                $\textbf{M}_{p}$ $\xrightarrow{}$ $\textbf{M}_{s}$ in \ourmethod{} & 8.03 (\textcolor{red}{+2.0}) & 4.49 (\textcolor{red}{+0.6}) & 4.95 (\textcolor{red}{+0.7})\\
               \bottomrule
    \end{tabular}
}
\end{table}

\noindent\textbf{The effect of the Model’s variants.}
Our method, \ourmethod{}, is built upon the vision-transformer model and is sensitive to different model configurations. 
To investigate the impact of varying transformer block settings, we conducted experiments with three distinct model configurations on the ISTD dataset. The results are summarized in Table~\ref{tab:variants}.
We observed that increasing the number of model parameters enhances shadow removal capabilities, leading to a significant improvement from 1.94MB to 2.52MB in overall image performance.
While higher model capacity improves performance, it also increases computational time, which can be burdensome.
To address this concern, we selected the middle model parameter configuration to strike a balance between performance and computational efficiency, alleviating the overhead.

\begin{table}[ht]
\centering
\caption{Comparisons of the Model's variants over ISTD dataset~\cite{wang2017stacked}. \label{tab:variants}}
\setlength{\tabcolsep}{0.4em}
\renewcommand{\arraystretch}{0.7}
\adjustbox{width=0.85 \linewidth}{
    \begin{tabular}{l|c|c|c|c}
        \toprule
         \multirow{2}{*}{Model Size} & Params  & {S}  & {NS}& {All}\\
         & (M) & MAE$\downarrow$  & MAE$\downarrow$ & MAE$\downarrow$ \\
                \midrule
                Small & 1.94 &6.61 & 3.93	&4.37	\\
                \midrule
               \rowcolor{Gray}  Middle & 2.28 & 6.08 & 3.86 & 4.23\\
                \midrule
                Large & 2.52 & 6.27 & 3.80 & 4.20\\
               \bottomrule
    \end{tabular}
}
\end{table}

\begin{figure}[t]
    \centering
    \includegraphics[width=.35\textwidth]{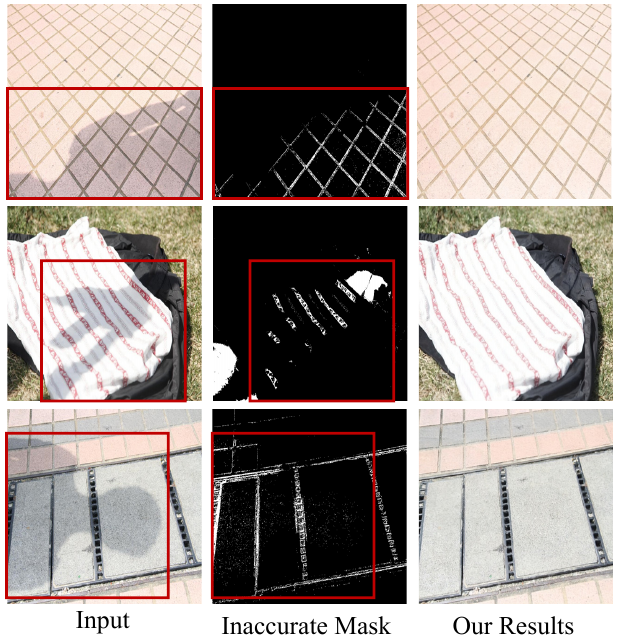} 
    \caption{\revision{Our shadow removal results with inaccurate masks from SRD dataset. From left to right: shadow image, inaccurate shadow mask, and the results using our proposed method. It is clearly shown that even with a highly inaccurate shadow mask, our method can still effectively remove the shadows, demonstrating the robustness of MAPE.}
    \label{fig:mask_srd} 
    }
\end{figure}

\begin{figure}[t]
    \centering
    \begin{subfigure}[b]{0.045\textwidth}
        \centering
        \includegraphics[width=\textwidth]{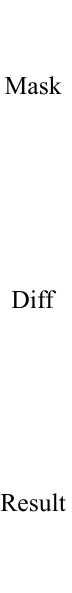}
    \end{subfigure}%
    \begin{subfigure}[b]{0.1083\textwidth}
        \centering
        \includegraphics[width=\textwidth]{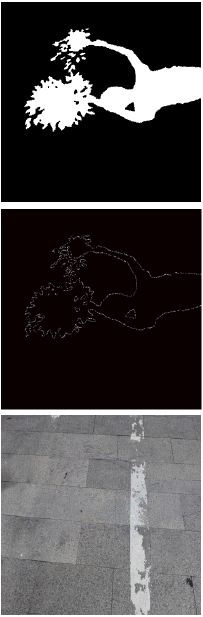}
        \caption{}
    \end{subfigure}%
    \begin{subfigure}[b]{0.108\textwidth}
        \centering
        \includegraphics[width=\textwidth]{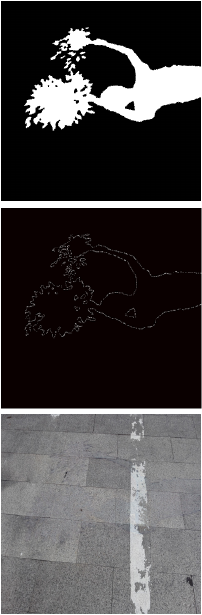}
        \caption{}
    \end{subfigure}%
    \begin{subfigure}[b]{0.1097\textwidth}
        \centering
        \includegraphics[width=\textwidth]{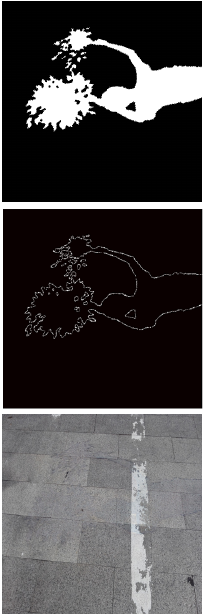}
        \caption{}
    \end{subfigure}%
    \begin{subfigure}[b]{0.1075\textwidth}
        \centering
        \includegraphics[width=\textwidth]{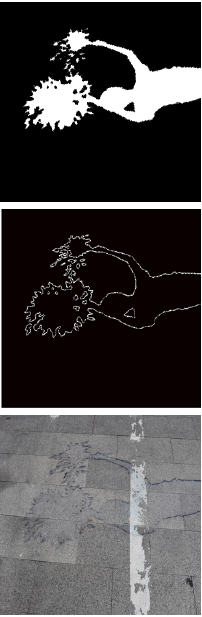}
        \caption{}
    \end{subfigure}
    \caption{\revision{(\emph{Top row}) Shadow masks with BER values of 0, 0.61, 1.82, and 4.39, corresponding to (a), (b), (c), and (d), respectively (where lower BER indicates higher accuracy). (\emph{Middle row}) Differences between the shadow masks from the ISTD dataset and the masks from the top row. (\emph{Bottom row}) Corresponding shadow removal results of our method.}}\label{fig:mask_compare}
\end{figure}

\noindent\textbf{\revision{The effect of the quality of masks.}}
\revision{Evaluating the robustness of MAPE to varying-quality shadow masks is crucial. We conducted a series of experiments and analyses to evaluate the robustness of MAPE to varying-quality shadow masks. Results suggest that MAPE maintains strong performance, even with inaccurate or low-quality masks.
Specifically, the SRD dataset~\cite{DeshadowNet2017} uses shadow masks predicted by DHAN~\cite{cun2019ghostfree}, which are incomplete and inaccurate, as shown in Figure~\ref{fig:mask_srd}. Despite this, our method achieves efficient shadow removal, demonstrating MAPE's robustness to inaccurate input masks.
To further quantify this robustness, we degraded the shadow masks in the ISTD dataset~\cite{wang2017stacked} to generate low-quality masks. Specifically, since the accuracy of shadow masks is related to the correctness of shadow boundaries, we first extracted the shadow region contours from the dataset's shadow masks. We then randomly misclassified contour shadow pixels as non-shadow pixels to generate low-quality masks.
As shown in the first row of Figure~\ref{fig:mask_compare}, different low-quality masks are presented, where the differences between low-quality and ISTD dataset masks are visible.
The quality of these masks was evaluated using the balance error rate (BER)~\cite{ber}, where lower BER indicates higher accuracy.
The results presented in Table~\ref{tab:mask-ber} indicate that shadow removal performance is closely related to mask quality. When the BER increases, reflecting lower mask quality, the shadow removal performance degrades. However, even when the BER increases sevenfold (from 0.61 to 4.39), the MAE only changes by 11\%, demonstrating that MAPE maintains its robustness to varying-quality masks. Finally, visual comparisons of shadow masks with different quality and the corresponding results are provided in Figure~\ref{fig:mask_compare}. We can observe slight contour issues in the shadow regions when BER is 0.61 or 1.82, and more pronounced issues when BER reaches 4.39.}

\begin{table}[ht]
\centering 
\caption{\revision{Impact of shadow mask quality on shadow regions (\textbf{S}), non-shadow regions (\textbf{NS}), and whole images (\textbf{All}) performance on ISTD dataset~\cite{wang2017stacked}. We evaluate six different mask quality levels, with BER used for assessment (lower BER indicates more accurate masks). The BER values range from 0.61 to 4.39. \label{tab:mask-ber}}}
\setlength{\tabcolsep}{0.4em}
\renewcommand{\arraystretch}{0.7}
\adjustbox{width=0.6 \linewidth}{
    \begin{tabular}{c|c|c|c}
        \toprule
        \revision{Mask} & \revision{S}  & \revision{NS}& \revision{All}\\
        \revision{BER $\downarrow$} & \revision{MAE$\downarrow$}  & \revision{MAE$\downarrow$} & \revision{MAE$\downarrow$} \\
                \midrule
                \revision{0.61} & \revision{7.13} & \revision{4.19} & \revision{4.67}\\
                \midrule
                \revision{1.29} & \revision{7.36} & \revision{4.21} & \revision{4.72}\\
                \midrule
                \revision{1.82} & \revision{7.78} & \revision{4.20} & \revision{4.78} \\
                \midrule
                \revision{2.85} & \revision{8.47} & \revision{4.22} & \revision{4.93}\\
                \midrule
                \revision{3.36} & \revision{8.87} & \revision{4.24} & \revision{5.02} \\
                \midrule
                \revision{4.39} & \revision{9.72} & \revision{4.29} & \revision{5.21}\\
               \bottomrule
    \end{tabular}
}
\end{table}

\begin{figure}[t]
    \centering
    \includegraphics[width=.45\textwidth]{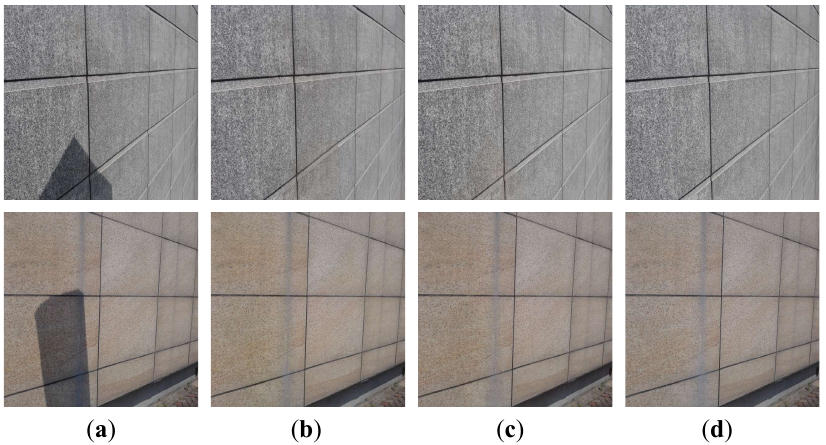} 
    \caption{\revision{Generalizability of ShadowMaskFormer on the unseen scenes.
    (a) The input shadow images from an unseen scenario. (b) Results of ShadowMaskFormer. (c) Results of ShdaowMaskFormer*. (d) The ground truth. As demonstrated, ShadowMaskFormer* effectively removes shadows even when the model has never encountered such scenes during training, highlighting its strong generalization capability on unseen scenes.}
    \label{fig:Generalizability} 
    }
\end{figure}

\noindent\textbf{\revision{Generalization Performance of ShadowMaskFormer.}} 
\revision{To assess the generalization capability of ShadowMaskFormer, we conducted experiments on previously unseen scenarios.
Specifically, we selected images from the ISTD dataset where shadows are cast on walls, an extremely common real-world scenario, as the unseen test set, while the remaining images were used for training.
We maintained the same experimental setup as described in Section~\ref{sec:setup}, ensuring consistency in the data partitioning strategy and evaluation approach.
Unlike artificial or synthetic test settings, wall shadows naturally occur in diverse environments with textured surfaces and varying reflectance properties, providing a representative benchmark for evaluating the generalizability of shadow removal models in practical applications.
The results, presented in Table~\ref{tab:Generalizability}, show that ShadowMaskFormer refers to the model trained and tested on the ISTD dataset, while ShadowMaskFormer* refers to the generalized results.
We observe that ShadowMaskFormer* shows a slight decrease in performance in the shadowed regions, while performance in non-shadow regions improves, with overall image performance remaining consistent. These findings suggest that our method effectively adapts to unseen shadow scenarios, handling novel shadow characteristics while maintaining robust global performance.}
\begin{table}[h]
\centering
\caption{\revision{Quantitative comparison of the proposed method on the ISTD dataset (ShadowMaskFormer) and generalization experiment (ShadowMaskFormer*) across shadow regions (\textbf{S}), non-shadow regions (\textbf{NS}), and whole images (\textbf{All}).}\label{tab:Generalizability}}
\setlength{\tabcolsep}{0.4em}
\renewcommand{\arraystretch}{0.7}
\adjustbox{width=0.65\linewidth}{
    \begin{tabular}{c|c|c|c}
        \toprule
        \multirow{2}{*}{\revision{Methods}} & \revision{S}  & \revision{NS}& \revision{All}\\
        & \revision{MAE$\downarrow$}  & \revision{MAE$\downarrow$} & \revision{MAE$\downarrow$} \\
                \midrule
                \revision{ShadowMaskFormer} & \revision{6.08}	& \revision{3.86}	& \revision{4.23}\\
                \midrule
                \revision{ShadowMaskFormer*} & \revision{6.69}	& \revision{3.55}	& \revision{4.23} \\
               \bottomrule
    \end{tabular}
}
\end{table}
Additionally, as shown in Figure~\ref{fig:Generalizability}, the visual results further demonstrate that our method successfully removes shadows even in previously unseen settings. This highlights the strong generalization capability of ShadowMaskFormer, enabling it to handle real-world shadow removal tasks effectively.

\section{Conclusion\label{sec:conclusion}}
In this paper, we introduce \ourmodel{}, a novel early mask utilization framework based on the transformer model, for efficient shadow removal. 
In \ourmodel{}, we successfully explore a simple and effective implementation of enhanced shadow region pixels during the patch embedding stage, enabling the model to restore shadow images. Specifically, after analyzing the characteristics of the previous methods, we develop the Mask Augmented patch embedding by using carefully the shadow mask. This approach efficiently encourages the model to explore the optimal shadow pixel gain factors as soon as possible. Experimental results on the ISTD, ISTD+, and SRD datasets demonstrate that our \ourmodel{} achieves outstanding performance compared to state-of-the-art methods by using fewer network parameters.
In the future, we will further explore more shadow removal methods based on the physical parameters of the shadow model, to enhance the interpretability of the model and its adaptability to different scenes.




\ifCLASSOPTIONcaptionsoff
  \newpage
\fi

\bibliography{egbib} 
\bibliographystyle{IEEEtran}

\end{document}